\documentclass[runningheads]{llncs}
\usepackage[T1]{fontenc}
\usepackage{graphicx}
\usepackage{microtype}
\usepackage{cite}
\usepackage{amsmath,amssymb,amsfonts}
\usepackage{algorithmic}
\usepackage{graphicx}
\usepackage{textcomp}
\usepackage{graphicx}
\usepackage{amsmath}
\usepackage{amssymb}
\usepackage{dsfont}
\usepackage{booktabs}
\usepackage[percent]{overpic}
\usepackage{makecell}
\usepackage{arydshln}
\usepackage[colorlinks=true, linkcolor=blue, citecolor=black, urlcolor=blue]{hyperref} % Enable hyperlinks with colors

\usepackage{colortbl}
\usepackage{multirow}
\usepackage{graphicx}
\usepackage{subcaption}  % For subfigures
\usepackage{caption}     % For captions
\usepackage{float}   
\DeclareMathOperator*{\argmin}{arg\,min}
\begin{document}
\title{A Study in Dataset Distillation for Image Super-Resolution}
\titlerunning{A Study in Dataset Distillation for Image Super-Resolution}
\author{Tobias Dietz\inst{1} \and
Brian B Moser\inst{1,2,3} \and
Tobias C Nauen\inst{1,2} \and
Federico Raue\inst{2} \and
Stanislav Frolov\inst{2} \and
Andreas Dengel\inst{1, 2}}
\authorrunning{Dietz et al.}
\institute{University of Kaiserslautern-Landau, Germany \and
German Research Center for Artificial Intelligence, Germany \and
Corresponding Author\\
\email{first.last@dfki}}
\maketitle              % typeset the header of the contribution
\begin{abstract}
Dataset distillation aims to compress large datasets into compact yet highly informative subsets that preserve the training behavior of the original data. While this concept has gained traction in classification, its potential for image Super-Resolution (SR) remains largely untapped. In this work, we conduct the first systematic study of dataset distillation for SR, evaluating both pixel- and latent-space formulations. We show that a distilled dataset, occupying only 8.88\% of the original size, can train SR models that retain nearly the same reconstruction fidelity as those trained on full datasets. Furthermore, we analyze how initialization strategies and distillation objectives affect efficiency, convergence, and visual quality. Our findings highlight the feasibility of SR dataset distillation and establish foundational insights for memory- and compute-efficient generative restoration models.

\end{abstract}

\section{Introduction}
Image super-resolution (SR) aims to reconstruct a high-resolution (HR) image from a low-resolution (LR) input. It is a key problem in computer vision, supporting applications in medical imaging, surveillance, and satellite analysis~\cite{moser2023hitchhiker, liu2022blind, moser2024diffusion,shen2025delt, xiao2025deep, vu2025comprehensive}. 
Modern SR models rely on large, high-quality datasets to learn fine-grained spatial structures. 
However, collecting, storing, and training on such datasets is costly and often impractical for memory- or compute-limited setups~\cite{ganguli2022predictability}. 
This dependency on large-scale data has become a central bottleneck in SR research and deployment.

Dataset distillation offers a potential solution. 
It aims to synthesize a compact set of representative samples that reproduce the training behavior of a much larger dataset~\cite{cazenavette2022dataset, cazenavette2023generalizing, aiello2024synthetic, yin2023squeeze}. 
By optimizing synthetic data to induce similar gradients or latent distributions as the original dataset, distillation can dramatically reduce data size while preserving model performance. 
Although this approach has achieved strong results in classification, its use in generative or pixel-to-pixel tasks like SR remains largely unexplored.

Applying dataset distillation to SR introduces new challenges. 
Unlike classification, which focuses on high-level semantic features, SR models must recover local textures, edges, and fine spatial patterns. 
Furthermore, SR is typically unsupervised (i.e., there are no discrete labels to guide synthesis), making gradient matching and feature alignment more complex. 
As a result, distilling an SR dataset requires capturing richer information than in standard supervised tasks.

In this paper, we take the first step toward bridging this gap. 
We perform a systematic study of dataset distillation for SR, adapting both pixel-space and latent-space techniques to evaluate their effectiveness for data-efficient SR training. 
We analyze the role of initialization, optimization objectives, and latent representations in maintaining reconstruction fidelity under extreme data reduction. 
Our experiments show that a dataset can be reduced by more than 91\% while preserving competitive SR quality. 
This work establishes a practical foundation for efficient SR training pipelines and opens a new direction for research in compact, generative data learning.

Our main contributions are:
\begin{itemize}
    \item We present the first systematic study of dataset distillation for image SR, establishing a new research direction.
    \item We compare pixel-space and latent-space approaches, analyzing how design choices such as initialization and pre-training affect performance.
    \item We demonstrate that SR datasets can be reduced by over 91\% while maintaining competitive reconstruction quality, highlighting the potential of dataset distillation for efficient SR training.
\end{itemize}

\section{Background}

\subsection{Image Super-Resolution}
Image super-resolution (SR) aims to recover a high-resolution (HR) image $\mathbf{y} \in \mathbb{R}^{s \cdot H \times s \cdot W \times C}$ from a single low-resolution (LR) input $\mathbf{x} \in \mathbb{R}^{H \times W \times C}$, where $s$ is the upscaling factor. 
The LR image is typically modeled as the result of a degradation process $D$, such that $\mathbf{x} = D(\mathbf{y})$. 
This process is often simplified to bicubic downsampling but can also include blur, noise, and compression artifacts. 
Because multiple HR images can map to the same LR image, SR is a fundamentally \textbf{ill-posed inverse problem}~\cite{moser2023hitchhiker}.

To address this, modern SR methods learn a mapping function $\psi_\theta: \mathbf{x} \mapsto \mathbf{y}$ parameterized by $\theta$ using deep neural networks. 
These models are trained on large datasets $\mathcal{T}$ of paired LR–HR samples, optimizing the parameters $\theta$ to minimize a reconstruction loss between the predicted HR image $\psi_\theta(\mathbf{x}_i)$ and the ground truth $\mathbf{y}_i$. 
The most common objective is the pixel-wise Mean Squared Error (MSE) or $L_2$ loss:
\begin{equation}
    \theta^{*} = \argmin_\theta \mathbb{E}_{(\mathbf{x}_i, \mathbf{y}_i) \in \mathcal{T}} \lVert \psi_\theta (\mathbf{x}_i) - \mathbf{y}_i \rVert^2.
\end{equation}
Despite their success, such models depend heavily on large-scale, high-quality datasets, making SR training resource-intensive.

\subsection{Dataset Distillation}
Dataset distillation aims to synthesize a compact set of synthetic samples $\mathcal{S}$ that reproduces the training behavior of a large dataset $\mathcal{T}$. 
Formally, given a real dataset $\mathcal{T} = \{(x_i, y_i)\}_{i=1}^N$, the goal is to generate a much smaller synthetic dataset $\mathcal{S}$ of size $M \ll N$. 
This is formulated as an optimization problem in which the synthetic data themselves are treated as trainable parameters:
\begin{equation}\small
    \mathcal{S}^* = \argmin_{\mathcal{S}} \mathcal{L}(\mathcal{S}, \mathcal{T}),
    \label{eq:def}
\end{equation}
where $\mathcal{L}$ denotes the distillation objective that aligns the learning dynamics of $\mathcal{S}$ and $\mathcal{T}$. 
For classification tasks, the size of $\mathcal{S}$ is often defined by the number of classes $\mathcal{C}$ and the desired number of images per class (IPC), such that $M = \mathcal{C} \cdot \text{IPC}$.

A key paradigm in dataset distillation is to match the gradients produced by real and synthetic data during training. 
The idea is that if both datasets induce similar gradient updates, a model trained on $\mathcal{S}$ will converge to a comparable state as one trained on $\mathcal{T}$. 

Dataset Condensation (DC)~\cite{cazenavette2022dataset} formalizes this concept by minimizing the cosine distance between gradients from synthetic and real batches. 
Given a loss function $\ell$, the DC objective is defined as:
\begin{equation}
\label{eq:dc}
    \mathcal{L}_{DC} (\mathcal{S}, \mathcal{T}) = 1 - 
    \frac{\nabla_\theta \ell^{\mathcal{S}}(\theta) \cdot \nabla_\theta \ell^{\mathcal{T}}(\theta)}
         {\|\nabla_\theta \ell^{\mathcal{S}}(\theta)\| \, \|\nabla_\theta \ell^{\mathcal{T}}(\theta)\|}.
\end{equation}
In its original form, $\ell$ is typically the cross-entropy loss used in classification. 
In this work, we adapt this formulation for image reconstruction tasks such as SR, where $\ell$ represents a pixel-level reconstruction loss.

\begin{figure*}[t!]
    \begin{center}
        \includegraphics[width=\textwidth]{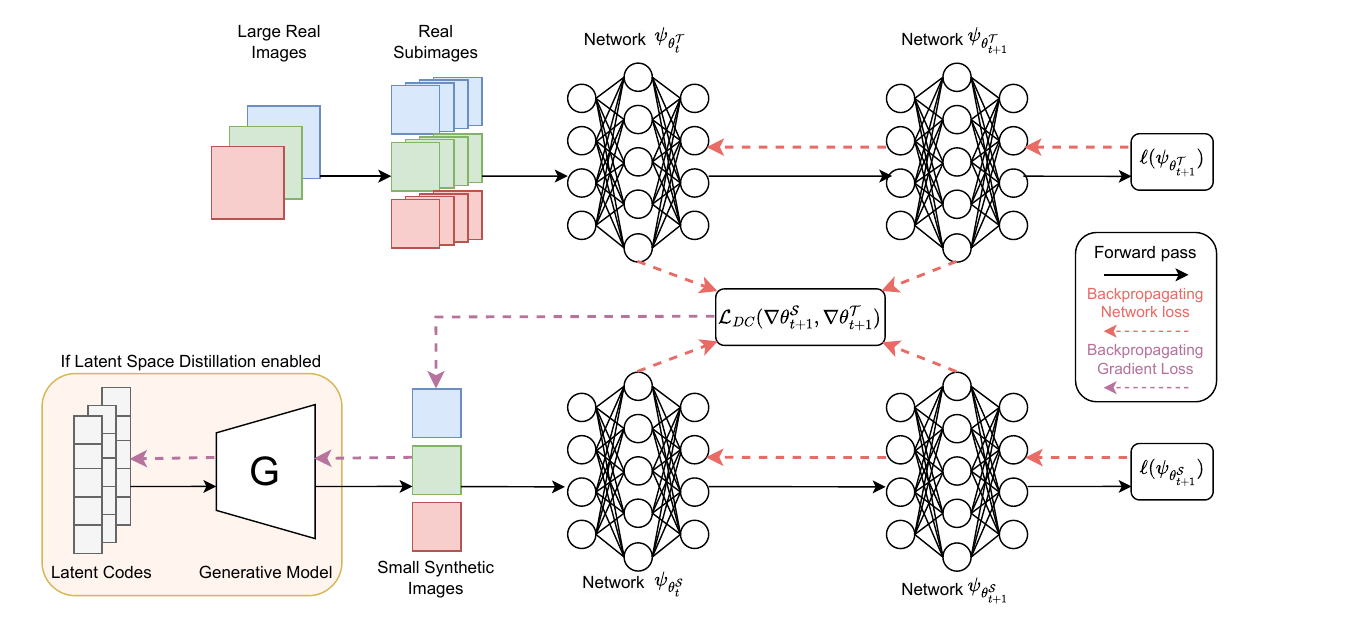}
        \caption{\label{fig:main}Overview of Dataset Condensation (DC) adapted for image SR. Our framework integrates a generative model to enable latent-space distillation and distills large SR datasets into smaller, highly representative synthetic samples.}
    \end{center} 
\end{figure*}

\section{Distilling Datasets for Super-Resolution}
\label{sec:methodology}

Our goal is to extend dataset distillation, originally designed for supervised classification, to the regression task of single-image super-resolution (SR). 
This adaptation introduces two main challenges. 
First, the standard distillation objective relies on gradient matching from a classification loss, which must be reformulated for pixel-wise image reconstruction. 
Second, SR datasets do not have discrete class labels, removing the structure that most distillation algorithms depend on.

To address these challenges, we propose three key adaptations. 
We first reformulate the gradient-matching objective for SR. 
Next, we introduce a pseudo-labeling scheme that enables class-conditional synthesis by grouping image patches. 
Finally, we extend distillation into the latent space of a generative model to improve efficiency and coherence.

\subsection{Gradient Matching for Image Reconstruction}
Our approach builds upon \textbf{Dataset Condensation (DC)}~\cite{zhao2020dataset}, which synthesizes a compact dataset $\mathcal{S}$ such that a model trained on $\mathcal{S}$ reaches similar performance to one trained on the full dataset $\mathcal{T}$. 
The core principle is to align the gradients of the training loss between real and synthetic data. 
For a network $\psi_\theta$, DC minimizes the distance between gradients computed on real samples $(\mathbf{x}_r, \mathbf{y}_r) \in \mathcal{T}$ and synthetic samples $(\mathbf{x}_s, \mathbf{y}_s) \in \mathcal{S}$.

Because SR is a regression problem, we replace the classification loss in DC with a reconstruction-based objective. 
We use the Mean Squared Error (MSE) between the predicted SR image $\psi_\theta(\mathbf{x})$ and the ground-truth high-resolution image $\mathbf{y}$:
\begin{equation}
    \ell_{SR}(\theta; \mathbf{x}, \mathbf{y}) = \|\psi_\theta(\mathbf{x}) - \mathbf{y}\|^2.
    \label{eq:sr_loss}
\end{equation}
Substituting this into the gradient-matching framework, we define the SR-specific distillation loss:
\begin{equation}
    \mathcal{L}_{SR-DC} = 
    \sum_{(\mathbf{x}_s, \mathbf{y}_s) \in \mathcal{S}} 
    D\!\left(
        \nabla_\theta \ell_{SR}(\theta; \mathbf{x}_s, \mathbf{y}_s),
        \nabla_\theta \ell_{SR}(\theta; \mathbf{x}_r, \mathbf{y}_r)
    \right),
\end{equation}
where $D$ is the cosine distance as in the original DC formulation (\autoref{eq:dc}). 
This objective trains the synthetic samples in $\mathcal{S}$ to encode the gradient information that drives effective SR training. 
An overview of the full framework is shown in \autoref{fig:main}.

\begin{figure*}[t!]
    \centering
    \includegraphics[width=.9\textwidth]{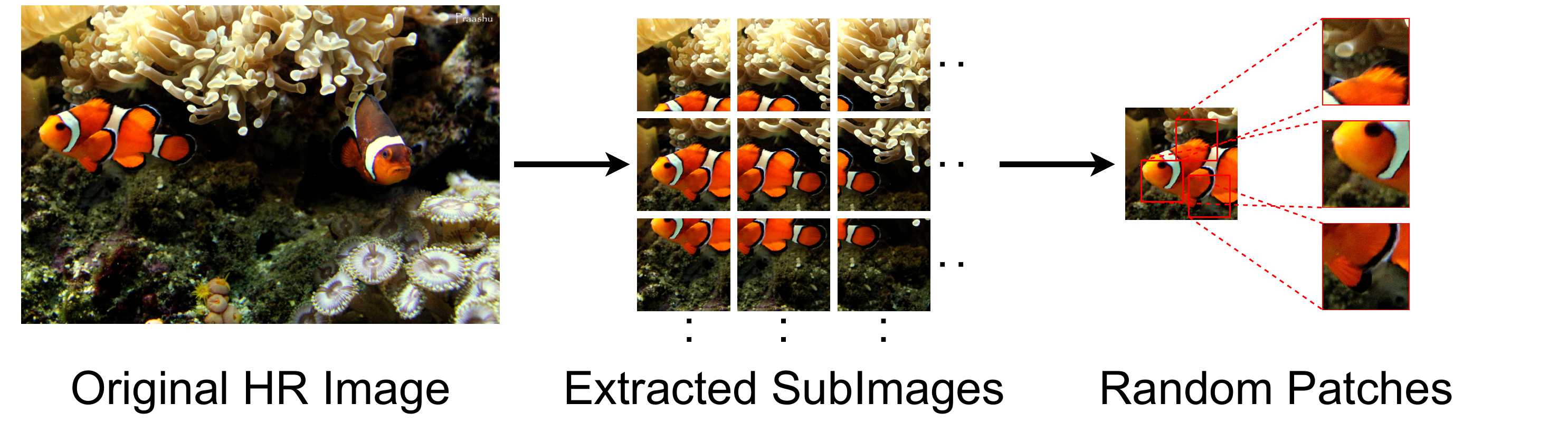}
    \caption{Illustration of the dataset preparation process. HR images are divided into overlapping sub-images using a stride to preserve edge details. During training, random patches are extracted from these sub-images to construct training batches. We will exploit this dataset preparation process to associate sub-images with their full image, effectively distilling the full size image into a sub-image.}
    \label{fig:datapreparation}
\end{figure*}

\subsection{Pseudo-Labels for Unsupervised Distillation}
A key obstacle in adapting distillation to SR is the absence of class labels. 
In classification, each class typically corresponds to one or more synthetic samples. 
Without this structure, SR distillation requires an alternative way to organize the data. 
We propose a simple but effective pseudo-labeling strategy that leverages the standard SR data preparation pipeline (\autoref{fig:datapreparation}).

High-resolution training images are first divided into overlapping sub-images (patches) to create training pairs. 
We treat all patches derived from the same original image as a single group and assign each group a unique pseudo-label. 
This grouping allows us to reinterpret SR distillation as a supervised task: 
for each pseudo-label (i.e., each HR image), we distill its many sub-images into a single, representative synthetic image. 

This design captures local and global structures efficiently while aligning with the class-based organization of standard distillation algorithms. 
In practice, this reduces approximately 32,000 sub-images to only 800 synthetic samples, a 91.12\% dataset reduction while retaining competitive reconstruction performance.

\subsection{Distillation in Latent Space}
Distilling directly in pixel space can be computationally demanding and often struggles to preserve long-range dependencies. 
To mitigate this, we perform distillation in the latent space of a pre-trained generative model, inspired by Generative Latent Distillation (GLaD)~\cite{cazenavette2023generalizing}. 
Here, the synthetic samples are represented by latent codes $\mathcal{Z} = \{z_1, \dots, z_M\}$, which are decoded into images through a generator $G$, such that $\mathbf{x}_s = G(z)$.

The optimization objective remains the same (matching gradients of the SR loss) but the optimization now updates the latent codes $z$ instead of raw pixels:
\begin{equation}
    \mathcal{L}_{Latent} = 
    D\!\left(
        \nabla_\theta \ell_{SR}(\theta; G(z_s), \mathbf{y}_s),
        \nabla_\theta \ell_{SR}(\theta; \mathbf{x}_r, \mathbf{y}_r)
    \right).
\end{equation}
This formulation reduces the dimensionality of the optimization problem and benefits from the strong image priors encoded in $G$. 
As a result, the distilled images are more coherent, visually consistent, and capable of preserving fine textures crucial for SR reconstruction.

\begin{table*}[t!]
\centering
\caption{Quantitative $2\times$ results using random initializations for dataset distillation with SRCNN. Best values for original data and synthetic data are
marked with bold letters, respectively.}
\label{table:noiseResultsSRCNN}
\resizebox{\columnwidth}{!}{
\begin{tabular}{cc  ccc ccc ccc}
                &     & \multicolumn{3}{c}{\textbf{Set5}}      & \multicolumn{3}{c}{\textbf{Set14}}     & \multicolumn{3}{c}{\textbf{DIV2K}}                                                \\ 
                \cmidrule(r){3-5} \cmidrule(r){6-8} \cmidrule(r){9-11}
                &     & \cellcolor[HTML]{C2C2C2}\textbf{PSNR} $\uparrow$ & \cellcolor[HTML]{C2C2C2}\textbf{SSIM} $\uparrow$            & \cellcolor[HTML]{C2C2C2}\textbf{LPIPS} $\downarrow$  & 
                        \cellcolor[HTML]{C2C2C2}\textbf{PSNR} $\uparrow$ & \cellcolor[HTML]{C2C2C2}\textbf{SSIM} $\uparrow$            & \cellcolor[HTML]{C2C2C2}\textbf{LPIPS} $\downarrow$  & 
                        \cellcolor[HTML]{C2C2C2}\textbf{PSNR} $\uparrow$    & \cellcolor[HTML]{C2C2C2}\textbf{SSIM} $\uparrow$            & \cellcolor[HTML]{C2C2C2}\textbf{LPIPS} $\downarrow$  \\ 
                        
    & \cellcolor[HTML]{C2C2C2}SRCNN & 34.8370   & 0.9394    & 0.0930    & \cellcolor[HTML]{EEEEEE}30.6087   & \cellcolor[HTML]{EEEEEE}0.8845    & \cellcolor[HTML]{EEEEEE}0.1529    & 33.3828   & 0.9269    & 0.1488 \\ 
    & \cellcolor[HTML]{C2C2C2}VDSR  & 32.0620   & 0.9127    & 0.1293    & \cellcolor[HTML]{EEEEEE}28.7188   & \cellcolor[HTML]{EEEEEE}0.8483    & \cellcolor[HTML]{EEEEEE}0.1896    & 31.2650   & 0.8970    & 0.1889 \\ 
\multirow{-3}{*}{\begin{tabular}[c]{@{}c@{}}Baseline\\(original)\end{tabular}}    
    & \cellcolor[HTML]{C2C2C2}EDSR  & \textbf{35.7720}      & \textbf{0.9442}   & \textbf{0.0886}   & \cellcolor[HTML]{EEEEEE}\textbf{31.5102}  & \cellcolor[HTML]{EEEEEE}\textbf{0.8937}   & \cellcolor[HTML]{EEEEEE}\textbf{0.1400} & \textbf{34.6383}     & \textbf{0.9374}   &  \textbf{0.1328}  \\ \hline
    
    & \cellcolor[HTML]{C2C2C2}SRCNN & \cellcolor[HTML]{EEEEEE}15.4093 & \cellcolor[HTML]{EEEEEE}0.4574 & \cellcolor[HTML]{EEEEEE}0.3176 & 14.9583 & 0.4890 & 0.3326 & \cellcolor[HTML]{EEEEEE}15.3261 & \cellcolor[HTML]{EEEEEE}0.4726  & \cellcolor[HTML]{EEEEEE}0.4151 \\
    & \cellcolor[HTML]{C2C2C2}VDSR  & \cellcolor[HTML]{EEEEEE}15.4176 & \cellcolor[HTML]{EEEEEE}0.4787 & \cellcolor[HTML]{EEEEEE}0.3709 & 14.9669 & 0.4739 & 0.3848 & \cellcolor[HTML]{EEEEEE}15.4559 & \cellcolor[HTML]{EEEEEE}0.4715  & \cellcolor[HTML]{EEEEEE}0.4587 \\ 
\multirow{-3}{*}{Syn. IPC=1}  & \cellcolor[HTML]{C2C2C2}EDSR  
    & \cellcolor[HTML]{EEEEEE}16.4814 & \cellcolor[HTML]{EEEEEE}\textbf{0.5132} & \cellcolor[HTML]{EEEEEE}0.4705 & 16.6783 & 0.4943 & 0.4714 & \cellcolor[HTML]{EEEEEE}17.8057 & \cellcolor[HTML]{EEEEEE}\textbf{0.5539} & \cellcolor[HTML]{EEEEEE}0.5375 \\ 
    
    & \cellcolor[HTML]{C2C2C2}SRCNN & 15.8993 & 0.4768 & \textbf{0.3110} & \cellcolor[HTML]{EEEEEE}15.5450 & \cellcolor[HTML]{EEEEEE}\textbf{0.5154} & \cellcolor[HTML]{EEEEEE}\textbf{0.3239} & 15.8124 & 0.4898 & \textbf{0.4088} \\
    & \cellcolor[HTML]{C2C2C2}VDSR  & 14.8428 & 0.4284 & 0.4090 & \cellcolor[HTML]{EEEEEE}13.4794 & \cellcolor[HTML]{EEEEEE}0.4116 & \cellcolor[HTML]{EEEEEE}0.4492 & 13.6287 & 0.4074 & 0.5181                         \\
\multirow{-3}{*}{Syn. IPC=10} & \cellcolor[HTML]{C2C2C2}EDSR  
    & \textbf{17.1089} & 0.4706 & 0.4519 & \cellcolor[HTML]{EEEEEE}\textbf{17.2534} & \cellcolor[HTML]{EEEEEE}0.4739 & \cellcolor[HTML]{EEEEEE}0.4601 & \textbf{18.2324} & 0.5005 & 0.5389   
    
\end{tabular}
}
\end{table*}

\begin{table*}[t]
\caption{Quantitative $2\times$ results using downscaled images to initialize dataset distillation with SRCNN. Best values for original data (downscaled) and synthetic data are marked with bold letters, respectively.}
\label{table:downscaledInitResults}
\resizebox{\textwidth}{!}{
\centering
\begin{tabular}{cc ccc ccc ccc}
\multicolumn{1}{c}{} & \multicolumn{1}{c}{} & \multicolumn{3}{c}{\textbf{Set5}} & \multicolumn{3}{c}{\textbf{Set14}} & \multicolumn{3}{c}{\textbf{DIV2K}} \\
\cmidrule(r){3-5} \cmidrule(r){6-8} \cmidrule(r){9-11}
\multicolumn{1}{c}{} & \multicolumn{1}{l|}{} & \multicolumn{1}{c}{\cellcolor[HTML]{B2B2B2}\textbf{PSNR} $\uparrow$} & \multicolumn{1}{c}{\cellcolor[HTML]{B2B2B2}\textbf{SSIM} $\uparrow$} & \multicolumn{1}{c}{\cellcolor[HTML]{B2B2B2}\textbf{LPIPS} $\downarrow$} & \cellcolor[HTML]{B2B2B2}\textbf{PSNR} $\uparrow$ & \cellcolor[HTML]{B2B2B2}\textbf{SSIM} $\uparrow$ & \cellcolor[HTML]{B2B2B2}\textbf{LPIPS} $\downarrow$ & \cellcolor[HTML]{B2B2B2}\textbf{PSNR} $\uparrow$ & \cellcolor[HTML]{B2B2B2}\textbf{SSIM} $\uparrow$ & \cellcolor[HTML]{B2B2B2}\textbf{LPIPS} $\downarrow$ \\ 

\multicolumn{1}{c}{} & \multicolumn{1}{c}{\cellcolor[HTML]{C2C2C2}SRCNN} & \multicolumn{1}{c}{34.8370} & \multicolumn{1}{c}{0.9394} & \multicolumn{1}{c}{0.0930} & \cellcolor[HTML]{EEEEEE}30.6087 & \cellcolor[HTML]{EEEEEE}0.8845 & \cellcolor[HTML]{EEEEEE}0.1529 & 33.3828 & 0.9269 & 0.1488 \\ 

\multicolumn{1}{c}{} & \multicolumn{1}{c}{\cellcolor[HTML]{C2C2C2}VDSR}  & \multicolumn{1}{c}{32.0620} & \multicolumn{1}{c}{0.9127} & \multicolumn{1}{c}{0.1293} & \cellcolor[HTML]{EEEEEE}28.7188 & \cellcolor[HTML]{EEEEEE}0.8483 & \cellcolor[HTML]{EEEEEE}0.1896 & 31.2650 & 0.8970 & 0.1889 \\ 

\multicolumn{1}{c}{\multirow{-3}{*}{\begin{tabular}[c]{@{}c@{}}Baseline\\(original)\end{tabular}}} & \cellcolor[HTML]{C2C2C2}EDSR & \textbf{35.7720} & \textbf{0.9442} & \textbf{0.0886} & \cellcolor[HTML]{EEEEEE}\textbf{31.5102} & \cellcolor[HTML]{EEEEEE}\textbf{0.8937} & \cellcolor[HTML]{EEEEEE}\textbf{0.1400} & \textbf{34.6383} & \textbf{0.9374} & \textbf{0.1328} \\ 

& \cellcolor[HTML]{C2C2C2}SRCNN & \cellcolor[HTML]{EEEEEE}34.7127 & \cellcolor[HTML]{EEEEEE}0.9385 & \cellcolor[HTML]{EEEEEE}0.0942 & 30.6389 & 0.8844 & 0.1532 & \cellcolor[HTML]{EEEEEE}33.3287 & \cellcolor[HTML]{EEEEEE}0.9262 & \cellcolor[HTML]{EEEEEE}0.1522 \\

& \cellcolor[HTML]{C2C2C2}VDSR & \cellcolor[HTML]{EEEEEE}32.0619 & \cellcolor[HTML]{EEEEEE}0.9127 & \cellcolor[HTML]{EEEEEE}0.1293 & 28.7188 & 0.8483 & 0.1896 & \cellcolor[HTML]{EEEEEE}31.2650 & \cellcolor[HTML]{EEEEEE}0.8970 & \cellcolor[HTML]{EEEEEE}0.1889 \\

\multirow{-3}{*}{\begin{tabular}[c]{@{}c@{}}Baseline\\(downscaled)\end{tabular}} & \cellcolor[HTML]{C2C2C2}EDSR & \cellcolor[HTML]{EEEEEE}35.2664 & \cellcolor[HTML]{EEEEEE}0.9421 & \cellcolor[HTML]{EEEEEE}0.0898 & 30.8937 & 0.8853 & 0.1482 & \cellcolor[HTML]{EEEEEE}33.8184 & \cellcolor[HTML]{EEEEEE}0.9312 & \cellcolor[HTML]{EEEEEE}0.1388 \\\hline 

& \cellcolor[HTML]{C2C2C2}SRCNN & \textbf{33.9813} & 0.9349 & 0.0973 & \cellcolor[HTML]{EEEEEE}\textbf{30.2628} & \cellcolor[HTML]{EEEEEE}0.8812 & \cellcolor[HTML]{EEEEEE}0.1561 & \textbf{32.8649} & 0.9230 & 0.1544 \\

& \cellcolor[HTML]{C2C2C2}VDSR & 33.6614 & \textbf{0.9361} & 0.1002 & \cellcolor[HTML]{EEEEEE}30.1338 & \cellcolor[HTML]{EEEEEE}\textbf{0.8832} & \cellcolor[HTML]{EEEEEE}0.1579 & 32.7802 & \textbf{0.9256} & 0.1525 \\

\multirow{-3}{*}{Syn. IPC=1} & \cellcolor[HTML]{C2C2C2}EDSR & 33.6500 & 0.9352 & \textbf{0.0920} & \cellcolor[HTML]{EEEEEE}29.9498 & \cellcolor[HTML]{EEEEEE}0.8803 & \cellcolor[HTML]{EEEEEE}\textbf{0.1526} & 32.6784 & 0.9235 & \textbf{0.1465}                                 
\end{tabular}
}
\end{table*}

\begin{table*}[t]
\caption{Quantitative $2\times$ results using a pre-trained SRCNN during dataset distillation. Best values for original data and synthetic data are
marked with bold letters, respectively.}
\label{table:downscaledInitPretrainedResults}
\resizebox{\columnwidth}{!}{
\centering
\begin{tabular}{cc ccc ccc ccc}
    & & \multicolumn{3}{c}{\textbf{Set5}} & \multicolumn{3}{c}{\textbf{Set14}} & \multicolumn{3}{c}{\textbf{DIV2K}} \\
    \cmidrule(r){3-5} \cmidrule(r){6-8} \cmidrule(r){9-11}
    & & \cellcolor[HTML]{B2B2B2}\textbf{PSNR} $\uparrow$ & \cellcolor[HTML]{B2B2B2}\textbf{SSIM} $\uparrow$ & \cellcolor[HTML]{B2B2B2}\textbf{LPIPS} $\downarrow$ & \cellcolor[HTML]{B2B2B2}\textbf{PSNR} $\uparrow$ & \cellcolor[HTML]{B2B2B2}\textbf{SSIM} $\uparrow$ & \cellcolor[HTML]{B2B2B2}\textbf{LPIPS} $\downarrow$ & \cellcolor[HTML]{B2B2B2}\textbf{PSNR} $\uparrow$ & \cellcolor[HTML]{B2B2B2}\textbf{SSIM} $\uparrow$ & \cellcolor[HTML]{B2B2B2}\textbf{LPIPS} $\downarrow$ \\

    & \cellcolor[HTML]{C2C2C2}SRCNN & 34.8370 & 0.9394 & 0.0930 & \cellcolor[HTML]{EEEEEE}30.6087 & \cellcolor[HTML]{EEEEEE}0.8845 & \cellcolor[HTML]{EEEEEE}0.1529 & 33.3828 & 0.9269 & 0.1488 \\
    & \cellcolor[HTML]{C2C2C2}VDSR  & 32.0620 & 0.9127 & 0.1293 & \cellcolor[HTML]{EEEEEE}28.7188 & \cellcolor[HTML]{EEEEEE}0.8483 & \cellcolor[HTML]{EEEEEE}0.1896 & 31.2650 & 0.8970 & 0.1889 \\
\multirow{-3}{*}{\begin{tabular}[c]{@{}c@{}}Baseline\\(original)\end{tabular}}
    & \cellcolor[HTML]{C2C2C2}EDSR  & \textbf{35.7720} & \textbf{0.9442} & \textbf{0.0886} & \cellcolor[HTML]{EEEEEE}\textbf{31.5102} & \cellcolor[HTML]{EEEEEE}\textbf{0.8937} & \cellcolor[HTML]{EEEEEE}\textbf{0.1400} & \textbf{34.6383} & \textbf{0.9374} & \textbf{0.1328} \\\hline

    & \cellcolor[HTML]{C2C2C2}SRCNN & \cellcolor[HTML]{EEEEEE}33.9813 & \cellcolor[HTML]{EEEEEE}0.9349 & \cellcolor[HTML]{EEEEEE}0.0973 & 30.2628 & 0.8812 & 0.1561 & \cellcolor[HTML]{EEEEEE}32.8649 & \cellcolor[HTML]{EEEEEE}0.9230 & \cellcolor[HTML]{EEEEEE}0.1544 \\
    & \cellcolor[HTML]{C2C2C2}VDSR  & \cellcolor[HTML]{EEEEEE}33.6614 & \cellcolor[HTML]{EEEEEE}\textbf{0.9361} & \cellcolor[HTML]{EEEEEE}0.1002 & 30.1338 & \textbf{0.8832} & 0.1579 & \cellcolor[HTML]{EEEEEE}32.7802 & \cellcolor[HTML]{EEEEEE}\textbf{0.9256} & \cellcolor[HTML]{EEEEEE}0.1525 \\ 
\multirow{-3}{*}{\begin{tabular}[c]{@{}c@{}}Syn. IPC=1\\(random)\end{tabular}}   
    & \cellcolor[HTML]{C2C2C2}EDSR  & \cellcolor[HTML]{EEEEEE}33.6500 & \cellcolor[HTML]{EEEEEE}0.9352 & \cellcolor[HTML]{EEEEEE}\textbf{0.0920} & 29.9498 & 0.8803 & \textbf{0.1526} & \cellcolor[HTML]{EEEEEE}32.6784 & \cellcolor[HTML]{EEEEEE}0.9235 & \cellcolor[HTML]{EEEEEE}\textbf{0.1465} \\

    & \cellcolor[HTML]{C2C2C2}SRCNN & 33.9001 & 0.9321  & 0.1026 & \cellcolor[HTML]{EEEEEE}30.1633 & \cellcolor[HTML]{EEEEEE}0.8795 & \cellcolor[HTML]{EEEEEE}0.1623 & 32.7271 & 0.9198 & 0.1591 \\
    & \cellcolor[HTML]{C2C2C2}VDSR  & \textbf{33.9951}  & 0.9336 & 0.0995  & \cellcolor[HTML]{EEEEEE}\textbf{30.4137} & \cellcolor[HTML]{EEEEEE}0.8818 & \cellcolor[HTML]{EEEEEE}0.1559 & \textbf{32.9882} & 0.9228 & 0.1534 \\
\multirow{-3}{*}{\begin{tabular}[c]{@{}c@{}}Syn. IPC=1\\(pre-trained)\end{tabular}}
    & \cellcolor[HTML]{C2C2C2}EDSR  & 33.9310 & 0.9332 & 0.0961 & \cellcolor[HTML]{EEEEEE}30.2509 & \cellcolor[HTML]{EEEEEE}0.8805 & \cellcolor[HTML]{EEEEEE}0.1530 & 32.8899  & 0.9212 & 0.1529 
    
\end{tabular}
}
\end{table*}

\section{Experiments}

\subsection{Setup}
For the reference SR model $\psi_\theta$ used during dataset distillation, we chose the lightweight SRCNN \cite{dong2015image} model (randomly initialized if not stated otherwise). 
As a baseline comparison, we test the SR models SRCNN \cite{dong2015image}, VDSR \cite{kim2016accurate}, and EDSR \cite{lim2017enhanced}, trained on the distilled images, also on the full dataset.
For testing, we used the classical datasets Set5, Set14, and DIV2K (validation set) \cite{moser2023hitchhiker}.

\subsection{Pixel-Space Results}

\textbf{Initialization Matters.}
We first analyze the effect of initialization on pixel-space dataset distillation. 
As shown in \autoref{table:noiseResultsSRCNN}, starting from random noise fails to produce meaningful synthetic samples. 
Models trained on such datasets show large performance drops across all benchmarks. 
With one image per class (IPC=1), PSNR decreases by more than 15~dB compared to the baseline, and even at IPC=10, gains remain marginal. 
These results indicate that random initialization provides no useful structural prior, causing the distillation process to converge to poor local minima.

\textbf{Downscaled Image Initialization.}
To provide a more informative starting point, we initialize synthetic images by downscaling real high-resolution samples. 
This structured initialization significantly improves the quality of the distilled datasets. 
As shown in \autoref{table:downscaledInitResults}, SR models trained on downscaled initializations achieve performance much closer to the full dataset than those trained on random noise. 
However, while this approach produces reasonable reconstructions, the distilled datasets do not consistently surpass the performance of the downscaled baseline itself. 
This suggests that the distillation process struggles to refine these images further, especially in recovering high-frequency details, emphasizing the need for improved gradient alignment in pixel space.

\begin{table*}[t]
\caption{Quantitative results of SR models trained on the GAN inversion baseline and a synthetic dataset distilled with latent space distillation for x2 upscaling. For latent distillation, latent codes from GAN inversion are used as initialization and further optimized using a pretrained SRCNN to match its gradients. We compare both evaluations with the original baseline. Best values for original data and synthetic data are respectively marked with bold letters.}
\label{table:latentdistillationResults}
\centering
\resizebox{\textwidth}{!}{
\begin{tabular}{cc ccc ccc ccc}
    & & \multicolumn{3}{c}{\textbf{Set5}}    & \multicolumn{3}{c}{\textbf{Set14}} & \multicolumn{3}{c}{\textbf{DIV2K}}                                                                     \\
    \cmidrule(r){3-5} \cmidrule(r){6-8} \cmidrule(r){9-11}
    & & \cellcolor[HTML]{B2B2B2}\textbf{PSNR} $\uparrow$ & \cellcolor[HTML]{B2B2B2}\textbf{SSIM} $\uparrow$   & \cellcolor[HTML]{B2B2B2}\textbf{LPIPS} $\downarrow$  
      & \cellcolor[HTML]{B2B2B2}\textbf{PSNR} $\uparrow$ & \cellcolor[HTML]{B2B2B2}\textbf{SSIM} $\uparrow$            & \cellcolor[HTML]{B2B2B2}\textbf{LPIPS} $\downarrow$          
      & \cellcolor[HTML]{B2B2B2}\textbf{PSNR} $\uparrow$    & \cellcolor[HTML]{B2B2B2}\textbf{SSIM}  $\uparrow$   & \cellcolor[HTML]{B2B2B2}\textbf{LPIPS} $\downarrow$  \\
      
    & \cellcolor[HTML]{C2C2C2}SRCNN & 34.8370   & 0.9394    & 0.0930    & \cellcolor[HTML]{EEEEEE}30.6087   & \cellcolor[HTML]{EEEEEE}0.8845    & \cellcolor[HTML]{EEEEEE}0.1529   & 33.3828   & 0.9269    & 0.1488    \\
    & \cellcolor[HTML]{C2C2C2}VDSR  & 32.0620   & 0.9127    & 0.1293    & \cellcolor[HTML]{EEEEEE}28.7188   & \cellcolor[HTML]{EEEEEE}0.8483    & \cellcolor[HTML]{EEEEEE}0.1896   & 31.2650   & 0.8970    & 0.1889    \\
\multirow{-3}{*}{\begin{tabular}[c]{@{}c@{}}Baseline \\ (original)\end{tabular}}     
    & \cellcolor[HTML]{C2C2C2}EDSR  & \textbf{35.7720} & \textbf{0.9442}   & \textbf{0.0886} & \cellcolor[HTML]{EEEEEE}\textbf{31.5102} & \cellcolor[HTML]{EEEEEE}\textbf{0.8937} & \cellcolor[HTML]{EEEEEE}\textbf{0.1400} & \textbf{34.6383}  & \textbf{0.9374}   & \textbf{0.1328}   \\ 
    
    & \cellcolor[HTML]{C2C2C2}SRCNN & \cellcolor[HTML]{EEEEEE}34.2725 & \cellcolor[HTML]{EEEEEE}0.9361 & \cellcolor[HTML]{EEEEEE}0.0966 & 30.2008 & 0.8800 & 0.1590 & \cellcolor[HTML]{EEEEEE}32.9720 & \cellcolor[HTML]{EEEEEE}0.9225 & \cellcolor[HTML]{EEEEEE}0.1560 \\
    & \cellcolor[HTML]{C2C2C2}VDSR  & \cellcolor[HTML]{EEEEEE}32.0619 & \cellcolor[HTML]{EEEEEE}0.9127 & \cellcolor[HTML]{EEEEEE}0.1293 & 28.7188 & 0.8483 & 0.1896 & \cellcolor[HTML]{EEEEEE}31.2650 & \cellcolor[HTML]{EEEEEE}0.8970 & \cellcolor[HTML]{EEEEEE}0.1889 \\
\multirow{-3}{*}{\begin{tabular}[c]{@{}c@{}}Baseline \\ (GAN inversion)\end{tabular}} 
    & \cellcolor[HTML]{C2C2C2}EDSR  & \cellcolor[HTML]{EEEEEE}34.9169 & \cellcolor[HTML]{EEEEEE}0.9399 & \cellcolor[HTML]{EEEEEE}0.0903 & 30.5112 & 0.8834 & 0.1477 & \cellcolor[HTML]{EEEEEE}33.4698 & \cellcolor[HTML]{EEEEEE}0.9272 & \cellcolor[HTML]{EEEEEE}0.1435 \\\hline
    
    & \cellcolor[HTML]{C2C2C2}SRCNN & 34.3706  & 0.9360  & 0.0941 & \cellcolor[HTML]{EEEEEE}30.3550 & \cellcolor[HTML]{EEEEEE}0.8817 & \cellcolor[HTML]{EEEEEE}0.1553 & 33.0254 & 0.9231 & 0.1534 \\
    & \cellcolor[HTML]{C2C2C2}VDSR  & 32.0621  & 0.9127  & 0.1293 & \cellcolor[HTML]{EEEEEE}28.7188 & \cellcolor[HTML]{EEEEEE}0.8483 & \cellcolor[HTML]{EEEEEE}0.1896 & 31.2650 & 0.8970 & 0.1889 \\
\multirow{-3}{*}{\begin{tabular}[c]{@{}c@{}} Latent \\Distillation \end{tabular}}   
    & \cellcolor[HTML]{C2C2C2}EDSR  & \textbf{34.9590} & \textbf{0.9399} & \textbf{0.0889} & \cellcolor[HTML]{EEEEEE}\textbf{30.7149} & \cellcolor[HTML]{EEEEEE}\textbf{0.8846} & \cellcolor[HTML]{EEEEEE}\textbf{0.1458} & \textbf{33.5222} & \textbf{0.9274} & \textbf{0.1429}               
\end{tabular}
}
\end{table*}

\begin{table*}[t]
\caption{Quantitative results of EDSR networks trained on synthetic datasets from previous datasets for x4 upscaling. We observe latent distillation to be consistently better, except for LPIPS on Set5. Best values for original data and synthetic data are marked with bold letters.}
\label{table:x4upscalingResults}
\centering
\resizebox{\textwidth}{!}{
\begin{tabular}{c ccc ccc ccc}
    & \multicolumn{3}{c}{\textbf{Set5}} & \multicolumn{3}{c}{\textbf{Set14}} & \multicolumn{3}{c}{\textbf{DIV2K}} \\
    \cmidrule(r){2-4} \cmidrule(r){5-7} \cmidrule(r){8-10}
    \multicolumn{1}{c}{}   & \cellcolor[HTML]{B2B2B2}\textbf{PSNR}   $\uparrow$& \cellcolor[HTML]{B2B2B2}\textbf{SSIM}  $\uparrow$& \cellcolor[HTML]{B2B2B2}\textbf{LPIPS}  $\downarrow$
                            & \cellcolor[HTML]{B2B2B2}\textbf{PSNR}  $\uparrow$ & \cellcolor[HTML]{B2B2B2}\textbf{SSIM}  $\uparrow$& \cellcolor[HTML]{B2B2B2}\textbf{LPIPS}  
                             $\downarrow$& \cellcolor[HTML]{B2B2B2}\textbf{PSNR}   $\uparrow$& \cellcolor[HTML]{B2B2B2}\textbf{SSIM}  $\uparrow$& \cellcolor[HTML]{B2B2B2}\textbf{LPIPS}  $\downarrow$  \\ 
    
    \multicolumn{1}{c}{\cellcolor[HTML]{B2B2B2}\begin{tabular}[c]{@{}c@{}}Baseline \\ (original)\end{tabular}} 
        & 30.0909  & 0.8640 & 0.2095 & \cellcolor[HTML]{EEEEEE}26.5687 & \cellcolor[HTML]{EEEEEE}0.7425 & \cellcolor[HTML]{EEEEEE}0.3059 & 28.9269 & 0.8175 & 0.3050 \\ 
        
    \multicolumn{1}{c}{\cellcolor[HTML]{C2C2C2}\begin{tabular}[c]{@{}c@{}}Baseline \\ (downscaled)\end{tabular}} 
        & \cellcolor[HTML]{EEEEEE}29.0669 & \cellcolor[HTML]{EEEEEE}0.8474 & \cellcolor[HTML]{EEEEEE}\textbf{0.2171} & 25.8882 & 0.7246 & 0.3204 & \cellcolor[HTML]{EEEEEE}28.0390 & \cellcolor[HTML]{EEEEEE}0.7994 & \cellcolor[HTML]{EEEEEE}0.3190 \\ 
        
    \multicolumn{1}{c}{\cellcolor[HTML]{B2B2B2}\begin{tabular}[c]{@{}c@{}}Baseline \\ (GAN inversion)\end{tabular}} 
        & 29.3426 & 0.8508 & 0.2181 & \cellcolor[HTML]{EEEEEE}26.0815 & \cellcolor[HTML]{EEEEEE}0.7278 & \cellcolor[HTML]{EEEEEE}0.3169 & 28.3216 & 0.8024 & 0.3161 \\\hline
        
    \multicolumn{1}{c}{\cellcolor[HTML]{C2C2C2}\begin{tabular}[c]{@{}c@{}}Syn. IPC=1\\ (downscaled)\end{tabular}} & \cellcolor[HTML]{EEEEEE}27.9465 & \cellcolor[HTML]{EEEEEE}0.8281 & \cellcolor[HTML]{EEEEEE}0.2435 & 25.2692 & 0.7093 & 0.3331 & \cellcolor[HTML]{EEEEEE}27.4346 & \cellcolor[HTML]{EEEEEE}0.7849 & \cellcolor[HTML]{EEEEEE}0.3285 \\
    
    \multicolumn{1}{c}{\cellcolor[HTML]{B2B2B2}\begin{tabular}[c]{@{}c@{}}Syn. IPC=1\\ (pre-trained)\\ IPC=1\end{tabular}} 
        & 27.4473 & 0.8098 & 0.2574 & \cellcolor[HTML]{EEEEEE}24.9738 & \cellcolor[HTML]{EEEEEE}0.6964 & \cellcolor[HTML]{EEEEEE}0.3405 & 27.2203 & 0.7752 & 0.3353 \\

    \multicolumn{1}{c}{\cellcolor[HTML]{C2C2C2}\begin{tabular}[c]{@{}c@{}}Syn. IPC=1\\ (latent)\end{tabular}}
        & \cellcolor[HTML]{EEEEEE}\textbf{29.4424} & \cellcolor[HTML]{EEEEEE}\textbf{0.8521} & \cellcolor[HTML]{EEEEEE}0.2188 & \textbf{26.1528} & \textbf{0.7307} & \textbf{0.3149} & \cellcolor[HTML]{EEEEEE}\textbf{28.3765} & \cellcolor[HTML]{EEEEEE}\textbf{0.8036} & \cellcolor[HTML]{EEEEEE}\textbf{0.3152}
    
\end{tabular}
}
\end{table*}

\textbf{Pre-Trained Network During Distillation.}
We next study the impact of using a pre-trained SR model during dataset distillation. 
The motivation is intuitive: a model that already captures rich image priors may guide the synthetic samples toward more informative and realistic structures. 
However, as summarized in \autoref{table:downscaledInitPretrainedResults}, the results are mixed. 
While PSNR improves when using a pre-trained model, SSIM and LPIPS scores remain superior when the network is trained from scratch. 
This inconsistency suggests that pre-trained models, although providing strong gradient guidance, may also overfit to previously learned features that limit their adaptability during distillation. 
Despite the ambiguity, we retain the pre-trained SRCNN setup for subsequent experiments to examine potential long-term benefits in latent-space distillation.

\subsection{Latent-Space Results}
We next extend dataset distillation into the latent domain using a pre-trained generative model. 
Unlike pixel-space distillation, which operates on 192×192 image patches, latent-space distillation is constrained by the generator’s native resolution. 
Here, we adopt StyleGAN-XL~\cite{sauer2022stylegan} to generate 512×512 synthetic sub-images. 
The goal shifts from optimizing raw pixel values to optimizing latent codes $\mathcal{Z} = \{z_i\}$ that encode high-level image representations~\cite{moser2024latent, cazenavette2023generalizing}.

We initialize these latent codes using GAN Inversion with Pivotal Tuning~\cite{roich2022pivotal}, which leverages the learned prior of a pre-trained StyleGAN-XL model trained on ImageNet. 
This initialization provides a meaningful starting point without requiring a dedicated SR-specific generator. 
Although GAN inversion does not yield perfect reconstructions, it offers a strong initialization while remaining computationally efficient by limiting inversion iterations.

As shown in \autoref{table:latentdistillationResults}, SR models trained on latent-distilled datasets—using both GLaD and DC formulations—consistently outperform those trained on direct GAN inversions or earlier pixel-space methods. 
These results confirm that latent-space distillation preserves essential image structures for SR training, bridging the gap between synthetic and real datasets. 
Notably, latent distillation surpasses the downscaled initialization baseline across multiple architectures, including SRCNN and VDSR, highlighting its effectiveness for generalizable data synthesis.

\textbf{From 2$\times$ to 4$\times$.}
Finally, we evaluate how distilled datasets generalize to different upscaling factors. 
All previous experiments were distilled for 2$\times$ upscaling. 
We now reuse the same synthetic datasets to train SR models for 4$\times$ upscaling, testing whether a single distillation scale can generalize across resolutions. 
If successful, this would confirm that one synthetic dataset can serve multiple SR scales.

The results in \autoref{table:x4upscalingResults} support this hypothesis. 
Latent-distilled datasets maintain strong performance even at higher scales, outperforming all baselines and other distillation variants, except for LPIPS on Set5. 
This finding indicates that latent-space distillation captures robust, scale-invariant features, eliminating the need for separate distillations at each target resolution.

\begin{figure*}[t!]
    \begin{center}
        \includegraphics[width=\textwidth]{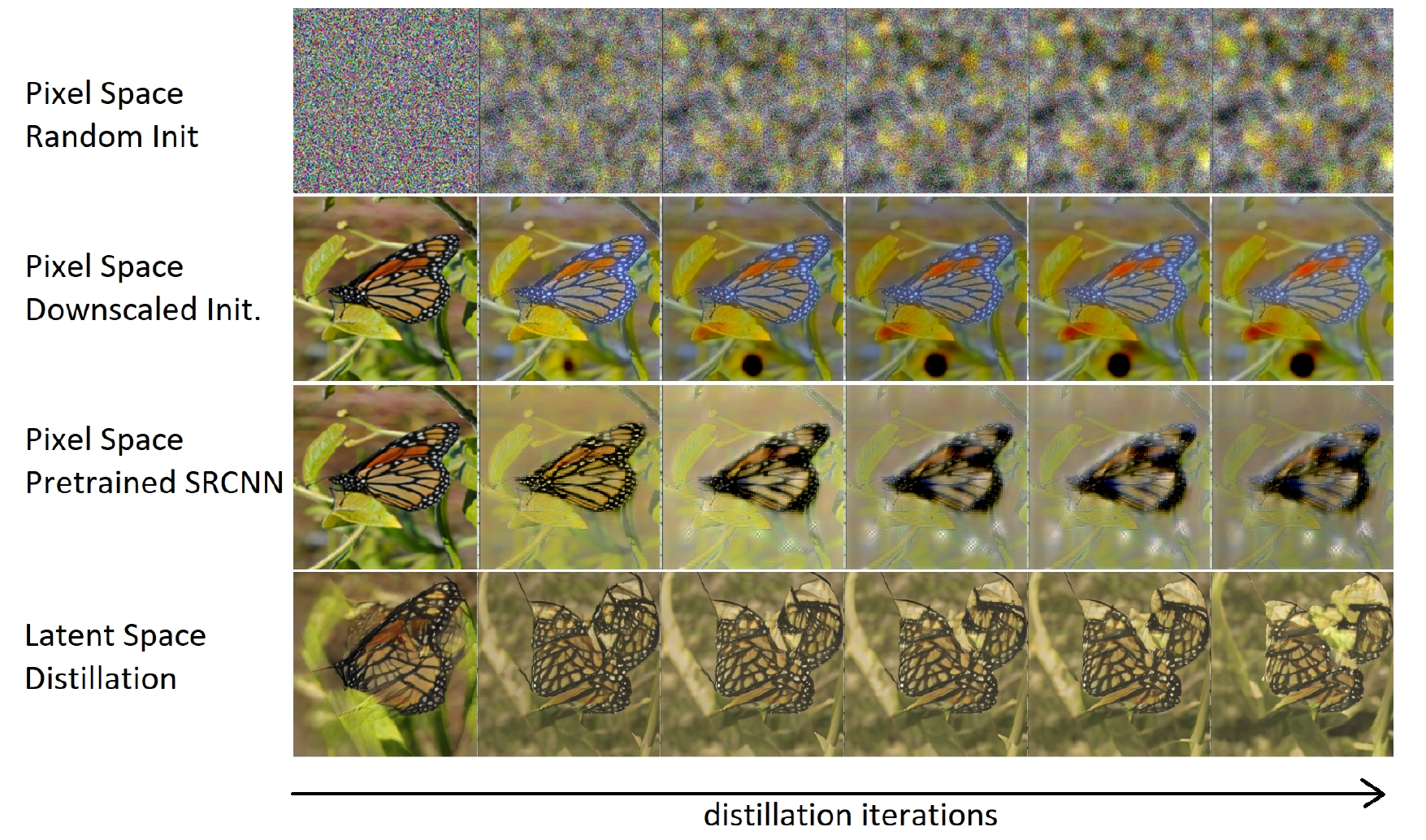}
        \caption{\label{fig:evolution}Illustrated evolution of synthetic high-resolution samples over distillation steps. Each row displays how the distillation process optimizes synthetic samples for different experiments over 1,000 distillation iterations. With the first row, it becomes clear that a good initialization is crucial for distilling SR images, as already quantified in our earlier experiments. Also, it becomes apparent that artifacts do not suddenly appear but develop gradually over distillation time. Furthermore, for latent distillation, only minor changes happen in later iterations, indicating that fewer iterations might suffice for a similarly well-performing synthetic dataset. }
    \end{center} 
\end{figure*}

\begin{figure*}[t!]
    \begin{center}
        \includegraphics[width=\textwidth]{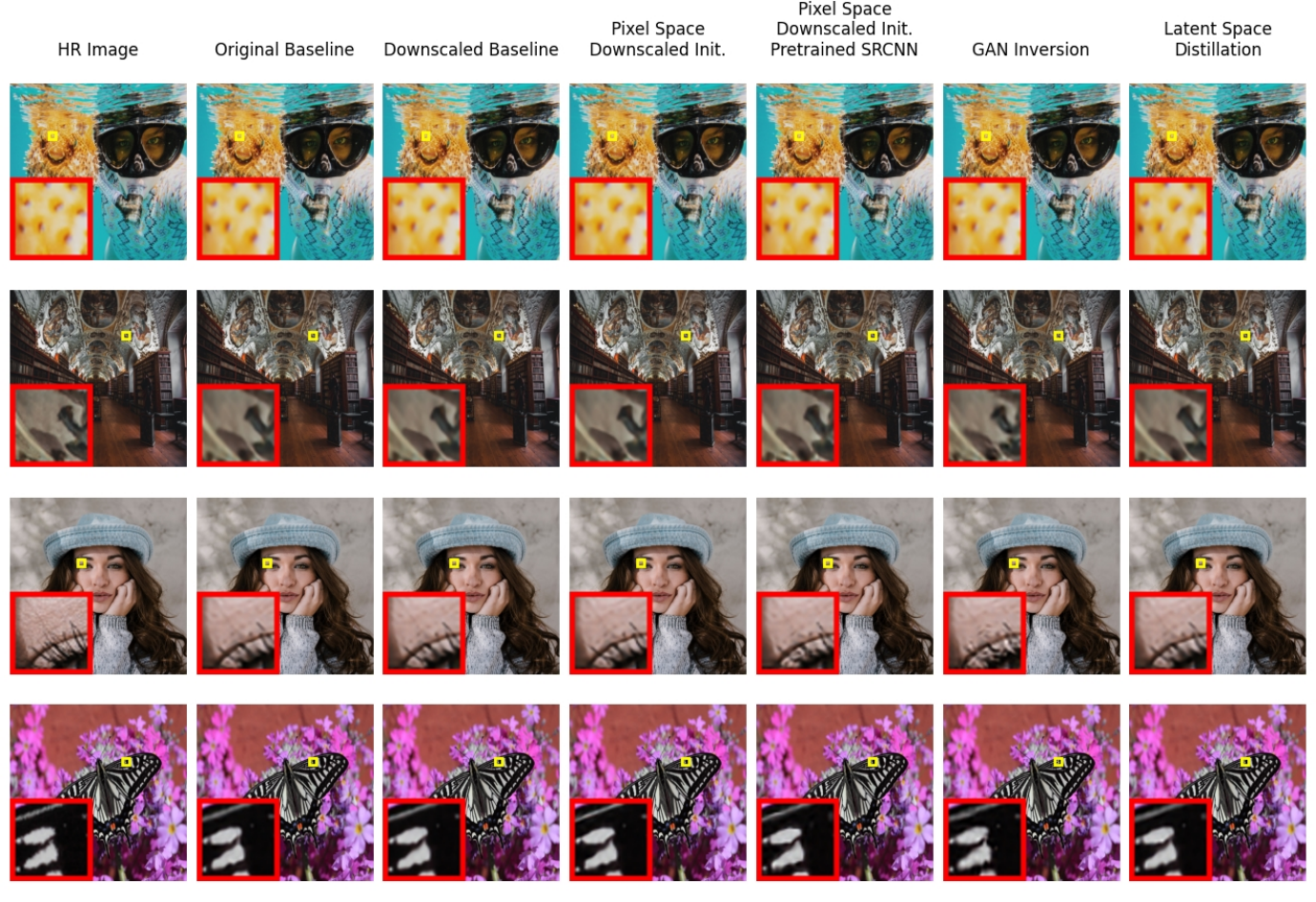}
        \caption{\label{fig:reconstructions}Comparison of qualitative results of the images generated by evaluating the different distillation experiments. Qualitatively, there is no significant difference between SR models trained on real or distilled images.}
    \end{center} 
\end{figure*}

\subsection{Qualitative Results}
Beyond quantitative metrics, we qualitatively examine both the distillation process and the visual quality of the resulting super-resolved images.

\subsubsection{Evolution of Distilled Samples.}
\autoref{fig:evolution} visualizes how latent-space distillation refines synthetic samples over time. 
In the fourth row, we observe a clear progression from initialization to the first synthetic output, driven by the structure of the generative model and its latent space. 
Subsequent iterations show only subtle refinements, suggesting that latent distillation converges faster than its pixel-space counterpart and may require fewer steps to achieve comparable performance. 
By contrast, pixel-space experiments (first three rows) exhibit little meaningful evolution. 
Artifacts accumulate gradually, indicating that these distortions are an inherent consequence of pixel-level optimization rather than undertraining.

\subsubsection{Analysis of Super-Resolution Outputs.}
For qualitative comparison, we visualize the outputs of EDSR models trained on different synthetic datasets. 
Results from pixel-space distillation with random initialization are omitted, as the corresponding models fail to reconstruct color and structural details. 
We focus instead on all other synthetic datasets, including the downscaled baseline, and compare them against models trained on real data using $4\times$ upscaling (differences at $2\times$ were negligible).

As illustrated in \autoref{fig:reconstructions}, even EDSR models trained on real data struggle to perfectly recover fine textures. 
However, models trained on distilled datasets produce reconstructions that are visually comparable. 
Minor deviations can be observed under close inspection, but no single distillation method clearly dominates based on qualitative appearance alone. 
In summary, although the distilled training images appear abstract and synthetic, the models trained on them successfully learn to reconstruct realistic textures in downstream SR tasks.

\section{Limitations and Future Work}
While our method removes the need for labeled data, it still depends on pseudo-label groupings to associate real and synthetic samples. 
This design stems from existing dataset distillation frameworks that optimize synthetic images per class, rather than jointly across all samples. 
Although this class-wise structure improves computational efficiency, it is not inherently aligned with SR, where the goal is to reconstruct fine spatial details rather than class-specific semantics. 
Future work should therefore explore class-agnostic or fully joint optimization schemes that preserve efficiency while encouraging greater diversity among distilled samples.

Another open question concerns the role of class-unconditional generative models in distillation. 
Exploring different backbones, such as unconditional GANs or Latent Diffusion Models, could further improve data diversity and reconstruction fidelity~\cite{moser2024latent}. 
We believe these directions will help uncover how generative priors influence the structure and generality of distilled data representations.

\section{Conclusion}
We have demonstrated that dataset distillation, traditionally developed for classification, can be successfully extended to the generative task of image Super-Resolution (SR). 
By reformulating gradient matching for pixel-to-pixel regression and introducing a pseudo-labeling mechanism to replace class supervision, we adapt the framework for unlabeled SR data. 
Through extensive experiments, we reveal a key distinction between approaches: while naive pixel-space distillation struggles to preserve image fidelity, latent-space distillation proves highly effective. 
By optimizing compact latent codes instead of raw pixels, we achieve over 91\% dataset reduction with only minor performance loss.

This work provides the first comprehensive study of dataset distillation for SR, establishing strong baselines and methodological insights. 
Our findings show that generative priors can serve as powerful enablers for data-efficient SR training and open a path toward scalable, memory-conscious learning for other pixel-based vision tasks. 
We hope this study inspires further exploration into distilling complex, continuous data distributions for broader applications in generative modeling and image restoration.

\section*{Acknowledgements}
This work was supported by the BMBF projects SustainML (Grant 101070408) and Albatross (Grant 01IW24002).

\bibliographystyle{splncs04}
\bibliography{refs}
\end{document}